\documentclass[runningheads]{llncs}
\usepackage{graphicx}
\usepackage{tabularx}
\usepackage{verbatim}

\usepackage{array}
\usepackage{booktabs}
\usepackage[acronym,nowarn]{glossaries}
\usepackage{multirow}
\usepackage{pdflscape}
\usepackage{pifont}
\usepackage{placeins}
\usepackage{siunitx}
\usepackage{subcaption}
\usepackage[dvipsnames,table,xcdraw]{xcolor}

\usepackage{url}
\usepackage[hidelinks]{hyperref}

\newcommand{\cmark}{\ding{51}}%
\newcommand{\xmark}{\ding{55}}%
\newcolumntype{x}[1]{%
>{\centering\arraybackslash\hspace{0pt}}p{#1}}%

\newacronym{ai}{AI}{Artificial Intelligence}
\newacronym{qa}{QA}{Question Answering}

\begin{document}
\title{Question rewriting? Assessing its importance for conversational question answering}
\titlerunning{Question rewriting? Assessing its importance for conversational QA}
%
\author{Gonçalo Raposo\orcidID{ 0000-0001-7806-6526} \and
Rui Ribeiro\orcidID{0000-0003-0922-1806} \and \\
Bruno Martins\orcidID{0000-0002-3856-2936} \and
Luísa Coheur\orcidID{0000-0002-2456-5028}}
\authorrunning{G. Raposo et al.}
%
\institute{INESC-ID, Instituto Superior Técnico, Universidade de Lisboa, Lisboa, Portugal
\email{\{goncalo.cascalho.raposo,rui.m.ribeiro,\\
bruno.g.martins,luisa.coheur\}@tecnico.ulisboa.pt}}
\maketitle              

\begin{abstract}

	\vspace*{-.84\baselineskip}
	In conversational question answering, systems must correctly interpret the interconnected interactions and generate knowledgeable answers, which may require the retrieval of relevant information from a background repository. Recent approaches to this problem leverage neural language models, although different alternatives can be considered in terms of modules for (a) representing user questions in context, (b) retrieving the relevant background information, and (c) generating the answer. This work presents a conversational question answering system designed specifically for the Search-Oriented Conversational AI (SCAI) shared task, and reports on a detailed analysis of its question rewriting module. In particular, we considered different variations of the question rewriting module to evaluate the influence on the subsequent components, and performed a careful analysis of the results obtained with the best system configuration. Our system achieved the best performance in the shared task and our analysis emphasizes the importance of the conversation context representation for the overall system performance.


\end{abstract}

\section{Introduction}
	
\begin{figure}[t]
        \centering
        \includegraphics[width=\textwidth]{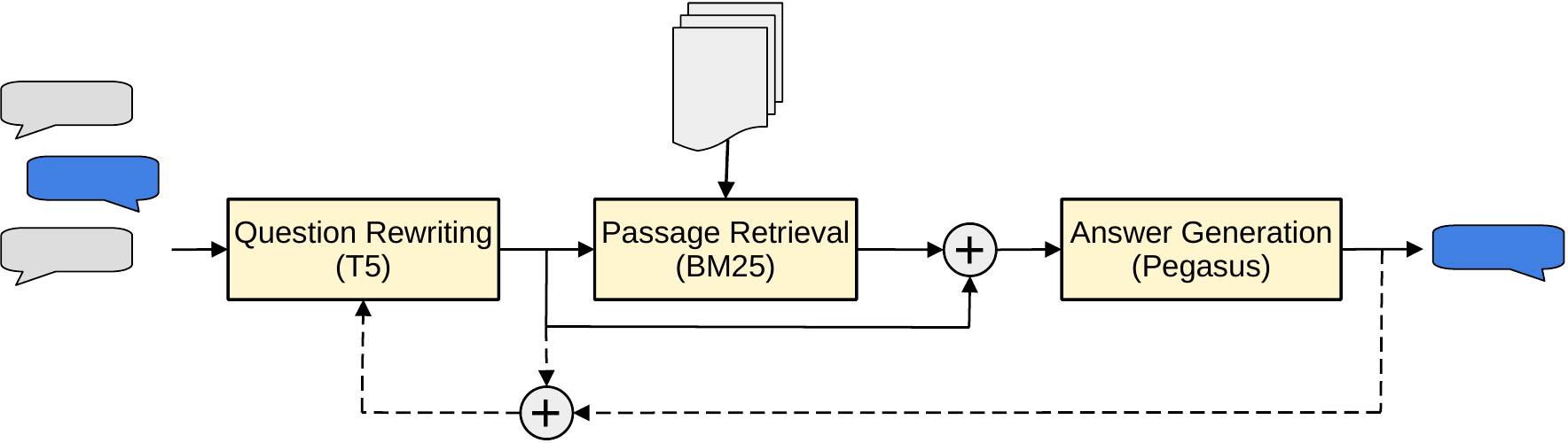}
        \caption{Proposed conversational question answering system. Question rewriting is performed using T5, passage retrieval using BM25, and answer generation using Pegasus. Dashed lines represent different inputs explored for question rewriting.}
        \label{fig:system}
        \vspace*{-\baselineskip}
    \end{figure}

	Conversational question answering extends traditional Question Answering (QA) by involving a sequence of interconnected questions and answers \cite{Choi2018}. Systems addressing this problem need to understand an entire conversation flow, often using explicit knowledge from an external datastore to generate a natural and correct answer for the given question. One way of approaching this problem is to divide it into 3 steps (see Fig. \ref{fig:system}): initial question rewriting, retrieval of relevant information regarding the question, and final answer generation.
	
	In a conversational scenario, questions may contain acronyms, coreferences, ellipses, and other natural language elements that make it difficult for a system to understand the question. Question rewriting aims to solve this problem by reformulating the question and making it independent of the conversation context \cite{Elgohary2019}, which has been shown to improve systems performance \cite{Vakulenko2021}.
	
	After an initial understanding of the question and its conversational context, the next challenge is the retrieval of relevant information to use explicitly in the answer generation \cite{Dalton2020}. For this step, the rewritten question is used as a query to an external datastore, and thus the performance of the initial rewriting module can affect the conversational passage retrieval \cite{Vakulenko2021a}.
	
	The last module has the task of generating an answer that incorporates the retrieved information conditioned on the rewritten question. The Question Rewriting in Conversational Context (QReCC) dataset \cite{Anantha2021} brings these tasks together, supporting the training and evaluation of neural models for conversational QA.
	
	This work presents a conversational QA system implemented according to the dataset and task definition of the Search-Oriented Conversational AI (SCAI) QReCC 2021 shared task\footnote{\url{https://scai.info/scai-qrecc/}}, specifically focusing on the question rewriting module. Besides evaluating the system performance as a whole, using many variations of the question rewriting module, our work highlights the importance of this module and how much it impacts the performance of  subsequent ones.
\section{Conversational Question Answering}
\label{section:cqa}

		To perform conversational question rewriting, the proposed system uses the model named \texttt{castorini/t5-base-canard}\footnote{\url{https://huggingface.co/castorini/t5-base-canard}} from HuggingFace \cite{Wolf2020}. This consists of a T5 model \cite{Raffel2020} which was fine-tuned for question rewriting using the CANARD dataset \cite{Elgohary2019}. No further fine-tuning was performed with QReCC data. 

        In order to incorporate relevant knowledge when answering the questions, our system uses a passage retrieval module built with Pyserini 
        \cite{Lin2021b}, i.e., an easy-to-use Python toolkit that allows searching over a document collection using sparse and dense representations. In our implementation, the relevant passage retrieval is performed using the BM25 ranking function \cite{Robertson2009}
        , with its parameters set to $k_1 = 0.82$ and $b=0.68$. This function is used to retrieve the top-10 most relevant passages.

        Since our system needs to extract the most important information from the retrieved passages, which are often large, we used a Transformer model pretrained for summarization. We chose the Pegasus model \cite{Zhang2020}, more specifically, the version \texttt{google/pegasus-large}\footnote{\url{https://huggingface.co/google/pegasus-large}}, which can handle inputs up to 1024 tokens. 
        
        We further fine-tuned the Pegasus model for 10 epochs in the task of answer generation, which can be seen as a summarization of the relevant text passages conditioned on the rewritten question. In detail, the training instances used the ground truth rewritten question concatenated with the ground truth passages (and additional ones retrieved with BM25), together with the ground truth answers as the target.

\section{Evaluation}

    \subsection{Experimental Setup}
    
        The dataset used for both training and evaluation was the one used in the SCAI QReCC 2021 shared task, which is a slight adaption of the QReCC dataset. The training data contains 11\,k conversations with 64\,k question-answer pairs, while the test data contains 3\,k conversations with 17\,k questions-answer pairs. For each question-answer pair, we have also the corresponding truth rewrites and passages, which are not considered during testing (unless specified otherwise).
        
        
        To evaluate each module, we used the same automatic metrics as the shared task: ROUGE1-R \cite{Lin2004} for question rewriting, Mean Reciprocal Rank (MRR) for passage retrieval, and F1 plus Exact Match (EM) \cite{Rajpurkar2016} for answer generation. We additionally used ROUGE-L for assessing answer generation.

    \subsection{Results}
    
        \subsubsection{Question Rewriting Input}
        
        We first studied different inputs to the question rewriting module in terms of the conversation history. Instead of using the original questions, one could replace them with the corresponding previous model rewrites. Moreover, one could use only the questions or also include the answers generated by the model. Regarding the length of the conversation history considered for question rewriting, we use all the most recent interactions that fit in the input size supported by the model. 

        \begin{table}[tb]
    \centering
    \vspace*{-\baselineskip}
    \caption{Evaluation of multiple variations of the input used in the question rewriting module: Question (Q), Model Answer (MA), Model Rewritten (MR).}
    \label{tab:results}
    \resizebox{\textwidth}{!}{%
        \begin{tabular}{@{}lrccccc@{}}
            \toprule
            \multirow{2}{*}{Description} & \multirow{2}{*}{Rewriting Input\hspace{0.4em}} & Rewriting & Retrieval & \multicolumn{3}{c}{Generation} \\ \cmidrule(l){3-3}\cmidrule(l){4-4} \cmidrule(l){5-7} 
             &  & ROUGE1-R & MRR & F1 & \hspace{1.3em}EM\hspace{1.3em} & ROUGEL-F1 \\ \midrule
            No rewriting ($h=1$) & \multicolumn{1}{c}{-} & 0.571 & 0.061 & 0.136 & 0.005 &  0.143 \\
            No rewriting ($h=7$) & \multicolumn{1}{c}{-} & 0.571 & 0.145 & 0.155 & 0.003 & 0.160 \\
            Questions & (Q) + Q & 0.673 & \textbf{0.158} & 0.179 & \textbf{0.011} & 0.181 \\
            Questions + answers & (Q + MA) + Q & 0.681 & 0.150 & 0.179 & 0.010 & 0.181 \\
            Rewritten questions & (MR) + Q & 0.676 & 0.157 & 0.187 & 0.010 & 0.188 \\
            Rewritten + answers & (MR + MA) + Q & \textbf{0.685} & 0.149 & \textbf{0.189} & 0.010 & \textbf{0.191} \\ \midrule
            Ground truth rewritten & \multicolumn{1}{c}{-} & 1 & 0.385 & 0.302 & 0.028 & 0.293 \\ \bottomrule
        \end{tabular}%
    }
    \vspace*{-.5\baselineskip}
\end{table}
        
        The results of our analysis are shown in Table \ref{tab:results}, where we observe that the system that did not perform question rewriting had the worst performance, especially when only the last question is considered ($h=1$). 
        
        When introducing question rewriting, we explored 4 variations of the question rewriting input, all exhibiting higher scores than without question rewriting. In particular, the highest scores occur in only 2 of them: when using only the questions and when using both the model rewritten questions and model answers. The variation where the system does not use model outputs in the question rewriting should be more resilient to diverging from the conversation topic.
        
        When we used the ground truth rewritten questions instead of performing question rewriting, the performance of the passage retrieval and answer generation components increased about $1.6 \sim 2.5 \times$, highlighting the importance of a good question rewriting.

        \subsubsection{Impact of Question Rewriting}
        After this initial evaluation, we used the system with the highest F1 score (rewriting using model rewritten questions and model answers) to further evaluate the impact of question rewriting. We computed the evaluation metrics for each sample and used the scores to classify the results into different splits reflecting result quality, allowing us to analyze a module's performance when the previous ones succeeded (\cmark) or failed (\xmark).
        
\begin{figure}[t]
            \centering
            \vspace*{-.25\baselineskip}
        	\begin{subfigure}[t]{0.49\textwidth}
                \centering
                \includegraphics[width=\textwidth]{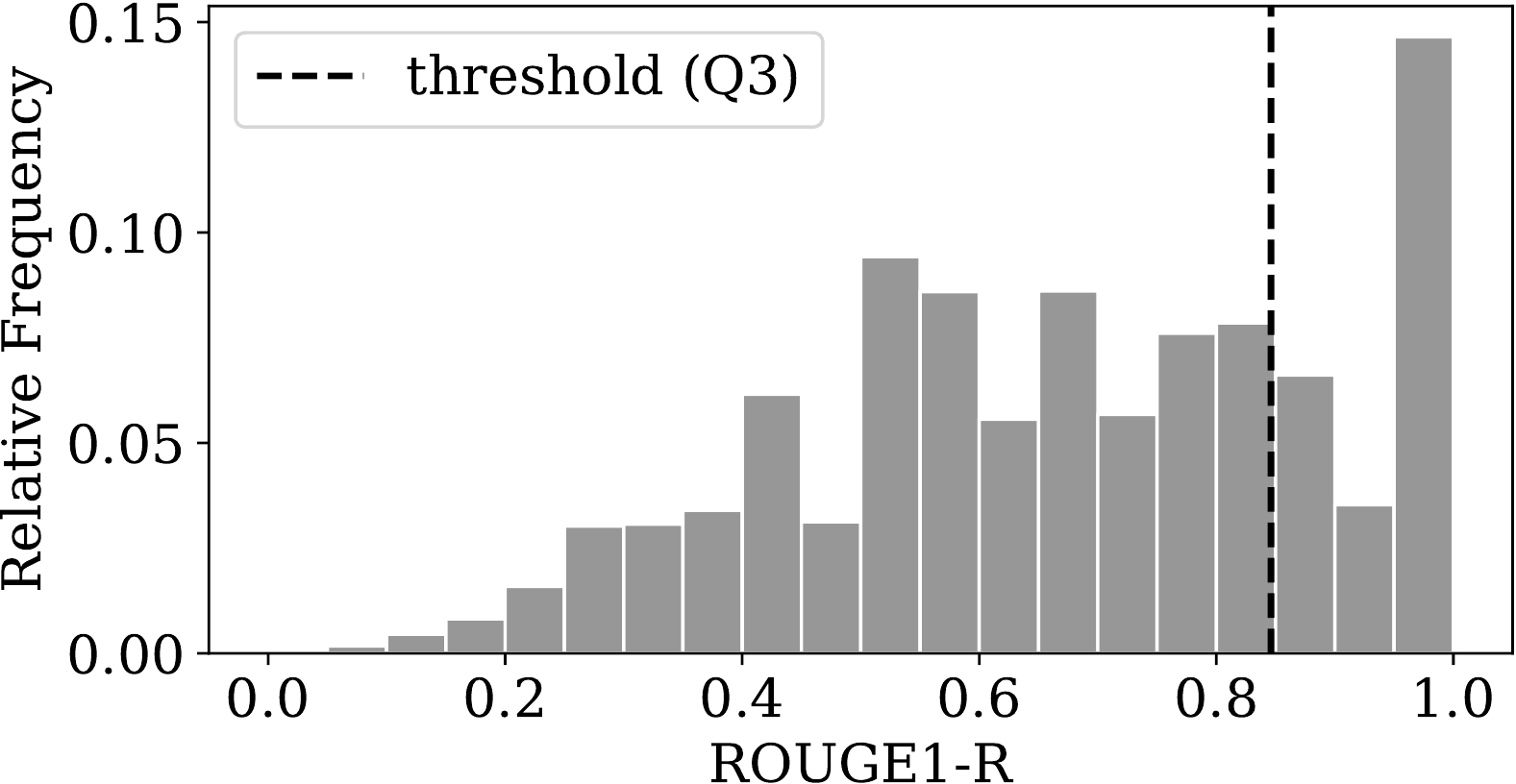}
                \caption{Distribution of ROUGE1-R scores for question rewriting.}
                \label{fig:rouge}
            \end{subfigure}
            \hfill
        	\begin{subfigure}[t]{0.49\textwidth}
                \centering
                \includegraphics[width=\textwidth]{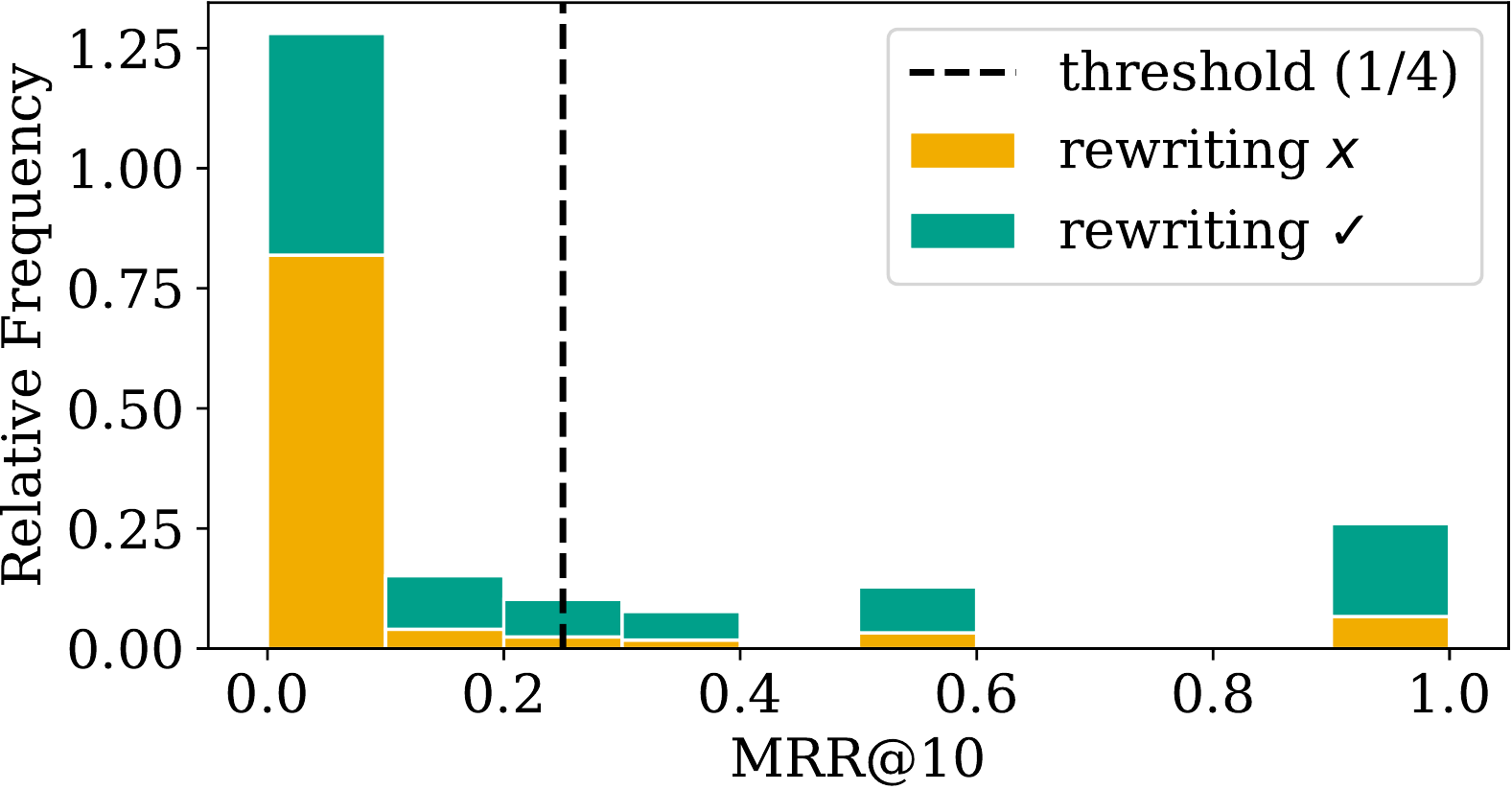}
                \caption{Distributions of MRR scores (retrieval) when question rewriting succeeds or fails.}
                \label{fig:mrr}
            \end{subfigure}
            \vspace*{-.25\baselineskip}
            \caption{Analysis of the influence of question rewriting on passage retrieval performance.  Relative frequencies refer to the number of samples of each split.} 
        \end{figure}

        To classify the performance of the question rewriting module using ROUGE scores, we used the 3\textsuperscript{rd} quartile of the score distribution as a threshold (shown in Fig. \ref{fig:rouge}), since we are unable to choose a value for an undoubtful classification. As for classifying the passage retrieval using the MRR score, an immediate option would be to classify values greater than 0 as successful. However, although our system retrieves the top-10 most relevant passages, the answer generation model is limited by its maximum input size, which resulted in less important passages being truncated. A preliminary analysis showed us that, in most samples, the model only considered $3 \sim 4$ passages, and therefore we defined the threshold of a successful retrieval as $\text{MRR} \geq 1/4$. 

        When the question rewriting succeeds ($\text{ROUGE1-R} \geq \mathrm{Q3}$), the passage retrieval also exhibits better performance, as seen by MRR scores greater than 0 being more than twice more frequent (see Fig \ref{fig:mrr}). Although both splits have many examples where the retrieval fails completely ($\text{MRR}=0$), they are about twice more frequent when the question rewriting fails.

        \begin{figure}[tb]
        	\centering
        	\begin{subfigure}[t]{0.49\textwidth}
        		\centering
        		\includegraphics[width=\textwidth]{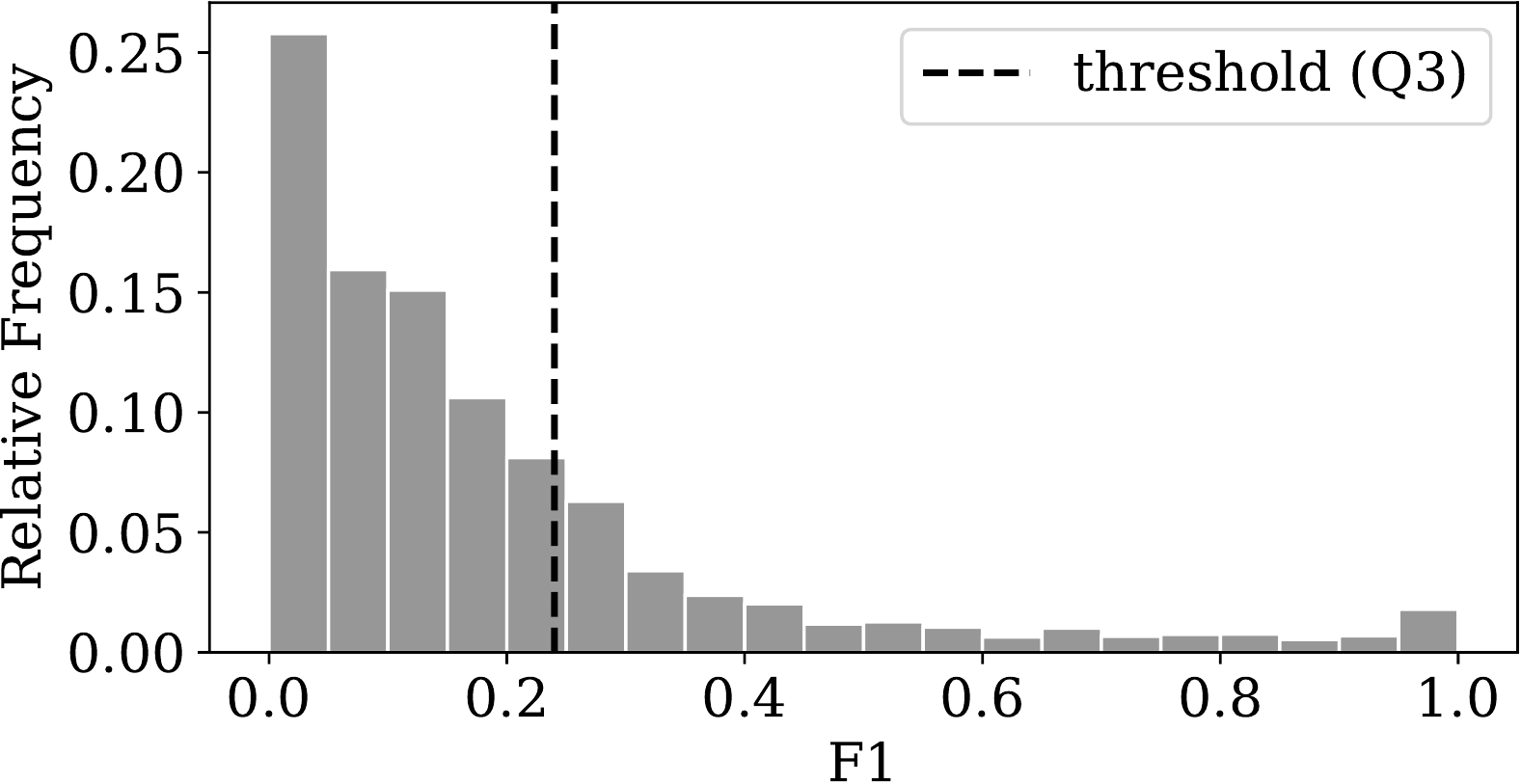}
        		\caption{Distribution of F1 scores for the answer generation component.}
        		\label{fig:f1}
        	\end{subfigure}
        	\hfill
        	\begin{subfigure}[t]{0.49\textwidth}
        		\centering
        		\includegraphics[width=\textwidth]{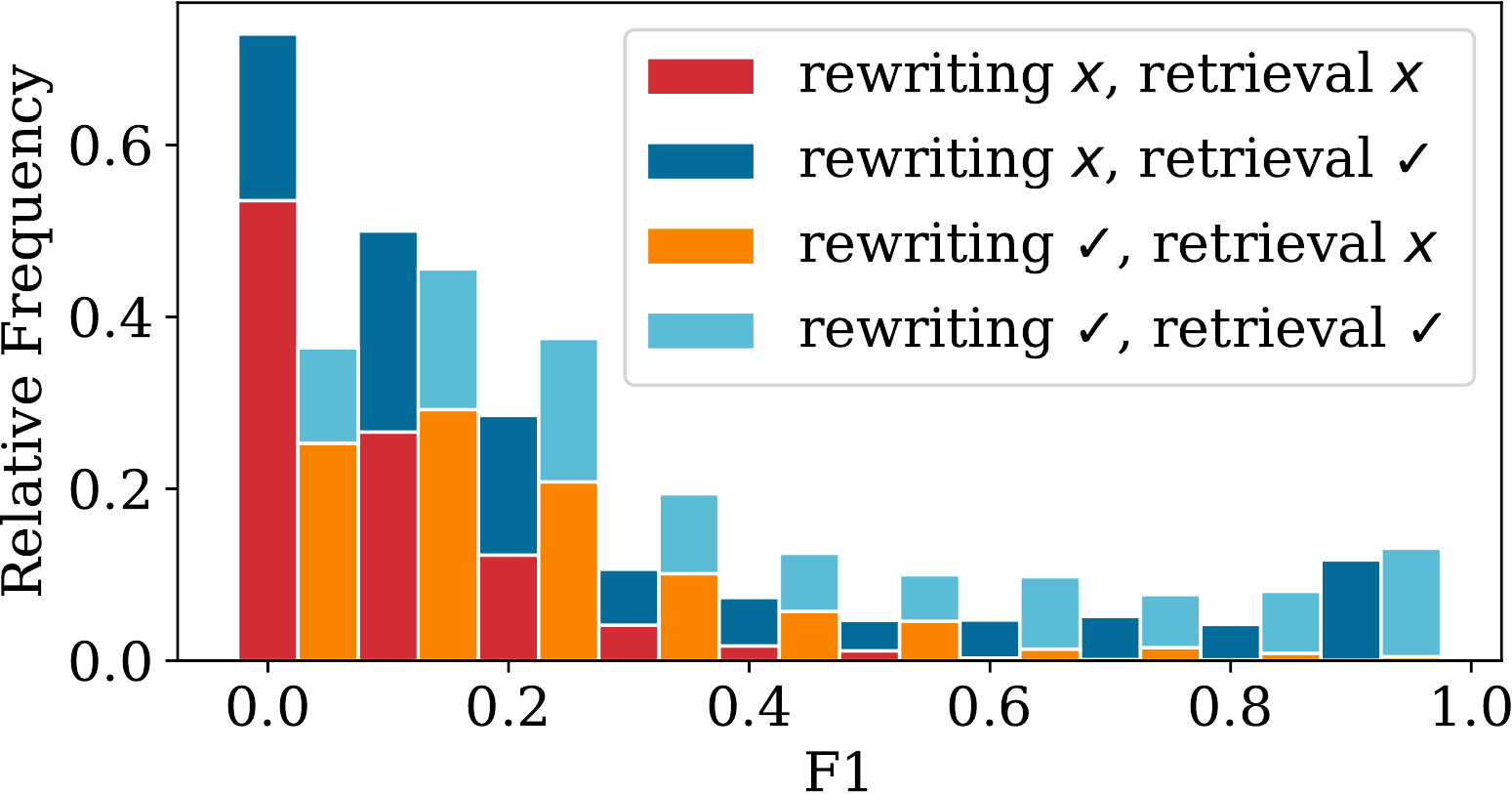}
        		\caption{Distributions of F1 scores (answer generation) when question rewriting and passage retrieval succeed and fail.}
        		\label{fig:f1_splits}
        	\end{subfigure}
        	\caption{Analysis of the influence of question rewriting and passage retrieval on answer generation performance. Relative frequencies refer to each split.} 
        	\vspace*{-1\baselineskip}
        \end{figure}

        Fig. \ref{fig:f1} presents the distribution of F1 scores for answer generation, showing that \SI{75}{\percent} of the results have an F1 score lower than 0.25. In turn, Fig. \ref{fig:f1_splits} shows 4 splits for when the question rewriting and retrieval modules each succeed or fail. Comparing the stacked bars together, one can analyze the influence of question rewriting in the obtained F1 score. Independently of the retrieval performance, F1 scores higher than 0.2 are much more frequent when the rewriting succeeds than when it fails. In particular, F1 scores between 0.3 and 0.8 are about $2\times$ more frequent when the rewriting succeeds. Moreover, poor rewriting performance results in about $2\times$ more results with an F1 score close to 0. Analyzing in terms of MRR, higher F1 scores are much more frequent when the retrieval succeeded. Interestingly, if the rewriting fails but the retrieval succeeds (less probable, as seen in Fig. \ref{fig:mrr}), the system is still able to generate answers with a high F1 score.

\subsubsection{Error Example}


In Table \ref{tab:error_examples}, we present a representative error where the system achieves a high ROUGE1-R score in the rewriting module but fails to retrieve the correct passage and to generate a correct answer. The only difference between the model and truth rewritten questions is in the omitted first name \textit{Ryan}, which led the system to retrieve a passage referring to a different person (\textit{Michael Dunn}).
Although the first name was not mentioned in the context, maybe by enhancing the question with information from the previous turn (e.g., the age or day of death) the system could have performed better in the subsequent modules.

Despite the importance of question rewriting, this example shows how a high ROUGE score in this module might not exactly reflect the ability to fully enhance the question with the necessary information from the conversation context.


\bgroup
\setlength{\tabcolsep}{3pt}
\def\arraystretch{1.1}
\begin{table}[!htb]
\centering
\vspace*{-1\baselineskip}
\caption{Example conversation where the retrieval and generation failed.}
\label{tab:error_examples}
\footnotesize
\begin{tabularx}{\textwidth}{c|c|X}
    \toprule
    \multicolumn{2}{c|}{Context} & Q: When was Dunn's death?  \\ 
    \multicolumn{2}{c|}{} & A: Dunn died on August 12, 1955, at the age of 59. \\
    \midrule
    \multicolumn{2}{c|}{Question} & What were the circumstances? \\ 
    \midrule
    \multirow{3}{*}{Rewriting} & Truth & What were the circumstances of Ryan Dunn's death? \\ \cline{2-3}
    & Model & What were the circumstances of Dunn's death? \\ \cline{2-3}
    & Score & \multicolumn{1}{l}{ROUGE1-R: 0.889} \\ 
    \midrule
    \multirow{3}{*}{Retrieval} & Truth & \url{http://web.archive.org/web/20191130012451id_/https://en.wikipedia.org/wiki/Ryan_Dunn_p3} \\ \cline{2-3}
    & Model & \url{https://frederickleatherman.wordpress.com/2014/02/16/racism-is-an-insane-delusion-about-people-of-color/?replytocom=257035_p1} \\ \cline{2-3}
    & Score & \multicolumn{1}{l}{MRR: 0} \\ 
    \midrule
    \multirow{3}{*}{Generation} & Truth & Ryan Dunn's Porsche 911 GT3 veered off the road, struck a tree, and burst into flames in West Goshen Township, Chester County, Pennsylvania. \\ \cline{2-3}
    & Model & The Florida Department of Law Enforcement concluded that Dunn's death was a homicide caused by a single gunshot wound to the chest. \\ \cline{2-3}
    & Score & \multicolumn{1}{l}{F1: 0.051, EM: 0, ROUGEL-F1: 0.128} \\
    \bottomrule
\end{tabularx}%
\end{table}
\egroup

\FloatBarrier
\section{Conclusions and Future Work}

    This work presented a conversational QA system composed of 3 modules: question rewriting, passage retrieval, and answer generation. The results obtained from its evaluation on the QReCC dataset show the influence of each individual module in the overall system performance, and emphasize the importance of question rewriting. When the question rewriting succeeded, both the retrieval and answer generation improved -- low scores were up to $2\times$ less frequent while higher scores were also about $2\times$ more frequent. 
    Future work should explore how to better control the question rewriting and its interaction with passage retrieval. Although our system with automatic question rewriting outperforms all the participants of the SCAI QReCC shared task, significant improvements can still be achieved with a better rewriting module. 

\subsubsection*{Acknowledgments}
	
	Work supported by national funds through Fundação para a Ciência e a Tecnologia (FCT), under project UIDB/50021/2020; by FEDER, Programa Operacional Regional de Lisboa, Agência Nacional de Inovação (ANI), and CMU Portugal, under project Ref. 045909 (MAIA) and research grant BI\textbar2020/090; and by European Union funds (Multi3Generation COST Action CA18231).


    
    \clearpage

\bibliographystyle{splncs04}
\bibliography{references.bib}

\begin{thebibliography}{10}
\providecommand{\url}[1]{\texttt{#1}}
\providecommand{\urlprefix}{URL }
\providecommand{\doi}[1]{https://doi.org/#1}

\bibitem{Anantha2021}
Anantha, R., Vakulenko, S., Tu, Z., Longpre, S., Pulman, S., Chappidi, S.:
  Open-domain question answering goes conversational via question rewriting.
  In: Proceedings of the 2021 Conference of the North American Chapter of the
  Association for Computational Linguistics: Human Language Technologies. pp.
  520--534. Association for Computational Linguistics, Online (Jun 2021).
  \doi{10.18653/v1/2021.naacl-main.44},
  \url{https://aclanthology.org/2021.naacl-main.44}

\bibitem{Choi2018}
Choi, E., He, H., Iyyer, M., Yatskar, M., tau Yih, W., Choi, Y., Liang, P.,
  Zettlemoyer, L.: Quac : Question answering in context (Aug 2018)

\bibitem{Dalton2020}
Dalton, J., Xiong, C., Callan, J.: Cast 2020: The conversational assistance
  track overview. In: The Twenty-Ninth Text REtrieval Conference(TREC 2020)
  Proceedings (2020), \url{https://trec.nist.gov/pubs/trec29/trec2020.html}

\bibitem{Elgohary2019}
Elgohary, A., Peskov, D., Boyd-Graber, J.: {Can You Unpack That? Learning to
  Rewrite Questions-in-Context}. In: {Proceedings of the 2019 Conference on
  Empirical Methods in Natural Language Processing and the 9th International
  Joint Conference on Natural Language Processing (EMNLP-IJCNLP)}. pp.
  5918--5924. Association for Computational Linguistics, Hong Kong, China (Nov
  2019). \doi{10.18653/v1/D19-1605}, \url{https://aclanthology.org/D19-1605}

\bibitem{Lin2004}
Lin, C.Y.: {ROUGE}: A package for automatic evaluation of summaries. In: Text
  Summarization Branches Out. pp. 74--81. Association for Computational
  Linguistics, Barcelona, Spain (Jul 2004),
  \url{https://aclanthology.org/W04-1013}

\bibitem{Lin2021b}
Lin, J., Ma, X., Lin, S.C., Yang, J.H., Pradeep, R., Nogueira, R.: Pyserini: A
  python toolkit for reproducible information retrieval research with sparse
  and dense representations. In: Proceedings of the 44th International ACM
  SIGIR Conference on Research and Development in Information Retrieval. p.
  2356–2362. SIGIR '21, Association for Computing Machinery, New York, NY,
  USA (2021). \doi{10.1145/3404835.3463238},
  \url{https://doi.org/10.1145/3404835.3463238}

\bibitem{Raffel2020}
Raffel, C., Shazeer, N., Roberts, A., Lee, K., Narang, S., Matena, M., Zhou,
  Y., Li, W., Liu, P.J.: Exploring the limits of transfer learning with a
  unified text-to-text transformer. Journal of Machine Learning Research
  \textbf{21}(140),  1--67 (2020), \url{http://jmlr.org/papers/v21/20-074.html}

\bibitem{Rajpurkar2016}
Rajpurkar, P., Zhang, J., Lopyrev, K., Liang, P.: Squad: 100, 000+ questions
  for machine comprehension of text. In: EMNLP (2016)

\bibitem{Robertson2009}
Robertson, S., Zaragoza, H.: The probabilistic relevance framework: Bm25 and
  beyond. Foundations and Trends® in Information Retrieval  \textbf{3}(4),
  333--389 (2009). \doi{10.1561/1500000019},
  \url{http://dx.doi.org/10.1561/1500000019}

\bibitem{Vakulenko2021}
Vakulenko, S., Longpre, S., Tu, Z., Anantha, R.: Question rewriting for
  conversational question answering. In: Proceedings of the 14th ACM
  International Conference on Web Search and Data Mining. p. 355–363. WSDM
  '21, Association for Computing Machinery, New York, NY, USA (2021).
  \doi{10.1145/3437963.3441748}, \url{https://doi.org/10.1145/3437963.3441748}

\bibitem{Vakulenko2021a}
Vakulenko, S., Voskarides, N., Tu, Z., Longpre, S.: A comparison of question
  rewriting methods for conversational passage retrieval (Jan 2021)

\bibitem{Wolf2020}
Wolf, T., Debut, L., Sanh, V., Chaumond, J., Delangue, C., Moi, A., Cistac, P.,
  Rault, T., Louf, R., Funtowicz, M., Davison, J., Shleifer, S., von Platen,
  P., Ma, C., Jernite, Y., Plu, J., Xu, C., {Le Scao}, T., Gugger, S., Drame,
  M., Lhoest, Q., Rush, A.: {Transformers: State-of-the-Art Natural Language
  Processing}. In: {Proceedings of the 2020 Conference on Empirical Methods in
  Natural Language Processing: System Demonstrations}. pp. 38--45. Association
  for Computational Linguistics, Online (October 2020).
  \doi{10.18653/v1/2020.emnlp-demos.6},
  \url{https://www.aclweb.org/anthology/2020.emnlp-demos.6}

\bibitem{Zhang2020}
Zhang, J., Zhao, Y., Saleh, M., Liu, P.: {PEGASUS}: Pre-training with extracted
  gap-sentences for abstractive summarization. In: III, H.D., Singh, A. (eds.)
  Proceedings of the 37th International Conference on Machine Learning.
  Proceedings of Machine Learning Research, vol.~119, pp. 11328--11339. PMLR
  (13--18 Jul 2020), \url{https://proceedings.mlr.press/v119/zhang20ae.html}

\end{thebibliography}

\end{document}